\documentclass[sigconf]{acmart}

\usepackage{caption}
\usepackage{subcaption}
\usepackage{multirow}
\usepackage{amsmath} 
\usepackage{bm}      

\AtBeginDocument{%
  \providecommand\BibTeX{{%
    \normalfont B\kern-0.5em{\scshape i\kern-0.25em b}\kern-0.8em\TeX}}}

\setcopyright{acmcopyright}
\copyrightyear{2021}
\acmYear{2021}
\acmDOI{}

\acmConference[KDD '21]{KDD '21: 27th ACM SIGKDD Conference on Knowledge Discovery and Data Mining}{August 14--18, 2021}{Virtual Event, Singapore}
\acmBooktitle{27th ACM SIGKDD Conference on Knowledge Discovery and Data Mining (KDD ’21), Machine Learning for Consumers and Markets Workshop, August 14--18, 2021, Virtual Event, Singapore}
\acmPrice{15.00}
\acmISBN{}

\begin{document}

\title{Uncovering the Source of Machine Bias}

\author{Xiyang Hu, Yan Huang, Beibei Li, Tian Lu}
\authornote{Names are in alphabetical order.}
\affiliation{%
  \institution{Carnegie Mellon University}
  \streetaddress{5000 Forbes Ave}
  \city{Pittsburgh}
  \state{PA}
  \country{USA}
  \postcode{15213}
}
\email{{xiyanghu, yanhuang, beibei.li}@cmu.edu, lutiansteven@gmail.com}

\renewcommand{\shortauthors}{Hu and Huang, et al.}

\begin{abstract}
We develop a structural econometric model to capture the decision dynamics of human evaluators on an online micro-lending platform, and estimate the model parameters using a real-world dataset. We find two types of biases in gender, i.e. \textit{preference-based bias} and \textit{belief-based bias}, are present in human evaluators' decisions. Both types of biases are in favor of female applicants. Through counterfactual simulations, we quantify the effect of gender bias on loan granting outcomes and the welfare of the company and the borrowers.
Our results imply that both the existence of the preference-based bias and that of the belief-based bias reduce the company's profits. When the preference-based bias is removed, the company earns more profits. When the belief-based bias is removed,  the company's profits also increase. Both increases result from raising the approval probability for borrowers, especially male borrowers, who eventually pay back loans.
For borrowers, the elimination of either bias decreases the gender gap of the true positive rates in the credit risk evaluation.
We also train machine learning algorithms on both the real-world data and the data from the counterfactual simulations. We compare the decisions made by those algorithms to see how evaluators' biases are inherited by the algorithms and reflected in machine-based decisions. We find that machine learning algorithms can mitigate both the preference-based bias and the belief-based bias.

\end{abstract}




\maketitle

\section{Introduction}

Humans and organizations often need to make decisions under imperfect information, and they generally rely on certain statistics that quantify the likelihood of different outcomes \citep{corbett2018measure}. However, in practice these statistics generally cannot perfectly predict the outcome of the event, and often involve human inputs, which may contain bias against certain demographics (often referred to as protected groups) such as minorities or females. As a result, members in these demographic groups are often treated unfairly, which leads to various social and economic problems.


A typical context of decision making under imperfect information is the loan approval decision in micro-lending. The emerging micro-lending business provides faster and convenient access to financial resources to more people with a streamlined loan application process \citep{mateescu2015peer}. The application process does not demand visits to a physical institution, and the credit assessment process involves human decisions \citep{lu2012social}. That is, human evaluators (i.e. platform staffs or lenders) make credit risk assessment with individual information (e.g., demographics) which is collected from loan applicants. Based on the evaluated credit risk, evaluators make the final loan approval decisions. However, this practice may contain bias due to limited human cognitive capacity to process complex computation for credit risk evaluation \citep{icard2018bayes}. Theoretically, bias in credit risk evaluation would heavily attenuate the evaluation accuracy, profitability of micro-lending platform or lenders, and cause social unfairness \citep{fu2019fair}. Unfortunately, the influence of human bias is largely neglected in literature. Since human evaluations are inevitable in many contexts such as micro-lending, the first goal of this study is to investigate: \textbf{Does bias exist in human evaluators' loan evaluation; if so, how does this bias affect their decisions and the market outcome?} 

Particularly, over the past decades, economists and social scientists have proposed different models to explain and quantify human bias \citep{bohren2019dynamics,arnold2018racial, gneezy2012toward, parsons2011strike, bertrand2005implicit, biernat1994shifting}. They classify human bias into two broad categories: \textbf{preference-based bias} (or tasted-based bias) and \textbf{belief-based bias} \citep{bohren2019dynamics}. Preference-based bias arises when evaluators have animus towards a particular group, while belief-based bias arises when evaluators' subjective beliefs about a group lead them to less favorably treat individuals from the group than members from other groups with the same observed performance. The classification of the two types of human bias is critical. It allows us to dynamically trace the evolution of bias during long-term (i.e. repeated) decisions rather than to regard bias as static. This static assumption of bias may overshadow the potentials and value of learning behaviors in decision making \citep{zhang2013forgetful}. It also enables us to unravel the influence of different types of human bias on decision outcomes.

Therefore, following such a classification of the source of the decision bias, in this study, we disentangle and quantify these two types of human bias in observable data from a micro-lending platform. In specific, we develop a structural econometric model of human evaluators' loan approval decisions, which captures the underlying economic processes that drive human evaluators' decisions. We then estimate the structural model based on the real-world data from a micro-lending platform that involves sufficient samples of repeated loan applications and approval decisions by the platform staff (i.e., human evaluators). Some of the parameters in the model capture the two possible types of human biases, and therefore, their parameter estimates can reveal whether human evaluators' decisions exhibit those two types of bias. We compare a number of alternative model specifications and identify the one that best explain the observed human decisions in the data. We also conduct a survey of human evaluators to confirm our chosen model specification. Using the estimated structural model, we conduct policy simulations to quantify the effects of these human bias and their welfare implications. 

Specifically, to measure the extent of bias, we examine the gender gaps for the approval true positive rate (TPR). We adopt the concept of equal opportunity, one of the most popular fairness notions. Equal opportunity suggests that qualified individuals, no matter what their sensitive attributes are, have equal opportunity to receive favorable outcomes \citep{hardt2016equality}. In our loan application setting, this means two gender group should have the same true positive rates if no gender bias exists.

Furthermore, with the recent boom of AI technologies, many organizations have increasingly adopted machine learning algorithms to replace human in decision making for better efficiency and decision quality. \footnote{The micro-lending platform we work with is also considering to use a machine learning algorithm to evaluate loan applications.} The original thought was algorithms are ``objective" and therefore have the potential to eliminate the discrimination against members in certain disadvantaged groups. However, numerous studies have cautioned that machine learning algorithms may also present non-negligible bias against those groups. 

One potential source of algorithm bias lies in some objective and inherent characteristics of data themselves. This type of cause has been well studied \citep{lu2019value, fuster2020predictably, kleinberg2018algorithmic, skeem2016risk}. For example, the correlations between the same features and the outcome may be different for different groups, and/or the data records of the disadvantaged group may be inferior in both quantity and quality than the regular group (the group that are not discriminated against historically). Another cause of algorithmic bias is the biased or flawed training data generated by humans. That is, human in the first place generates bias in the training data, which then translates into machine-learning assisted decision making \citep{fu2021crowds, corbett2018measure, tolan2019machine, hardt2016equality}. 

Therefore, in the second part of this paper, we intend to address the following question: \textbf{To what extent a machine learning algorithm trained on historical data inherit human bias?}  
To this end, we train various machines based on (a) data from human evaluators' decisions (the approved loans observed in the data) and (b) data from our counterfactual simulations with human biases removed. In the counterfactual simulation, we take advantage of an experiment the platform conducted: during a period of time, all applicants are approved without any selection. This novel experiment allows us to observe the credit behaviors of \textit{all} the applicants on the micro-lending platform which are usually unobservable.This allows us to infer the outcomes of the loans that are not approved by human evaluators, which usually go unobserved and are impossible to evaluate. By leveraging these "de-biased" data sets, we are able to compare machines trained based on data with and without human bias and demonstrate how different types of human bias, as well as other data characteristics, affect decisions made by machines.

We find that the evaluators exhibit both types of biases discussed above, with both preference-based bias and belief-based bias in favor of females. This is due to females generally behave better than males in repaying loans, and study has shown that females are more risk averse and ethically sensitive \citep{cumming2015gender}.  Although preference-based bias persists, belief-based bias gradually reduces as human evaluators learn from the repayment behaviors of each specific borrower.
Our policy simulations suggest that  the elimination of either the preference-based bias or the belief-based bias from human evaluators' decisions can increase the platform's profits.  
The mechanisms behind the two scenarios are the same. The extra profits result from lowering the approval probability of defaulters, especially female defaulters.
On the borrower end, the elimination of the two types of bias can mitigate the gender gap in the credit evaluation measured by the true positive rate . And when both types of bias are removed at the same time, the gender gap is the minimized.
When we further feed the real-world data and the counterfactual data into machine learning models, we find that eliminating upstream human decision biases help combat algorithmic biases.


The theoretical contribution of this study is multi-fold. First, while a great body of machine bias literature has focused on the bias generated in algorithm process (Suresh and Guttag 2020), our study is among the first to investigate the influence of human bias on AI-assisted decisions. Second, we introduce two types of human biases into a real-world decision-making context, and we are among the first to empirically identify and quantify these two human biases with a large-scale dataset and a structural econometric model. Furthermore, we also add to the micro-lending and FinTech literature by revealing how human bias could influence platforms' profitability and service equality. We also characterize how human evaluators make credit risk evaluations and loan approval decisions in practice. Methodologically, we design a novel and comprehensive empirical framework to uncover the behavioral sources of bias, which are  usually implicit and difficult to identify accurately. The framework works on secondary datasets, and enables to conduct counterfactual simulations, and AI-based analyses.

\section{Relevant Literature}

Our work is closely related to the literature on the bias and discrimination in human decision making, especially bias in financial evaluations.
Two categories of human decision biases have been extensively studied; one is ethnic or racial discrimination, another is gender discrimination. It has been documented that in financial loan market, both non-Fintech and Fintech lenders tend to discriminate against ethnic-minority borrowers (higher interest rates and lower probability of being funded) through liability document and facial bias \citep{sydnor2011sa, bartlett2019consumer}. It has also been that women are more credit constrained than men by microfinance institutions \citep{blanchflower2003discrimination}. In P2P lending, female borrowers need pay higher interest rates  \citep{alesina2013women, chen2017gender, chen2020gender}. And female founders are less successful attracting male investors compared to observably similar male founders \citep{ewens2020early}.

In addition to race bias and gender bias, other kinds of bias or discrimination common to see in society include immigration and age related bias \citep{dobbie2018measuring}, occupational related bias \citep{cui2019occupational}, home bias \citep{lin2016home}, etc. The major reasons behind these human decision bias lie in the minority applicants’ relative quality compared to the majority \citep{ferguson1995constitutes}, the decision makers' improper task objectives and incentives \citep{dobbie2018measuring}, and the inherent bias formation and evolution process of preference-based and belief-based evaluation biases \citep{gneezy2012toward, bohren2019dynamics}.

Economic theories classify inherent human bias into two types: preference-based bias and belief-based bias \citep{bohren2019dynamics}. Preference-based bias arise when evaluators have animus towards a particular group, while belief-based bias arise when evaluators' subjective beliefs about a group lead them to less favorably treat individuals from the group than members from the regular group with the same observed performance. Belief-based bias can be further classified into two subcategories: belief-based bias with misspecified/incorrect beliefs and belief-based bias with correct beliefs (sometimes referred to as statistical bias). The former occurs when the evaluators' subjective beliefs about the group-level statistics of the protected group are not the same as the reality, and the latter occurs when the subjective beliefs match the reality. In a static setting, preference and belief-based biases mix up with each other, and it is hard to disentangle their effects on human decisions. However, in a dynamic setting, we are able to distinguish between the two biases. This is because across periods, preference-based bias persists, while belief-based bias can be mitigated or even reversed when the evaluator observes new signals about each individual they are evaluating. In this paper, we follow the definitions of the two types of bias. Our context of multi-period microloan borrowing and lending provides a great setting to identify these two types of bias. 

Our paper also builds on the growing literature on algorithmic discrimination and machine learning bias.
One important source of machine bias is the human decision bias encoded in the training dataset.
\cite{fuster2020predictably} incorporate the prediction of machine learning models into a simple equilibrium model of finance credit, and find that algorithms increase rate disparity among and within different racial groups. \cite{stevenson2019algorithmic} conduct simulations to evaluate the race and age disparities in finance risk assessment, and demonstrate that human and machine interaction can lead to bias in both race and age. \cite{lambrecht2019algorithmic} find that economic forces in the market can distort neutral algorithms into discriminating females in terms of their exposure to advertisements of STEM (science, technology, engineering and mathematics) jobs.
Major reasons for algorithmic bias include lack of necessary data control (statistical bias) and unintended correlation with sensitive factors \citep{fu2019fair, bartlett2019consumer}, training-sample bias \citep{cowgill2020biased}, market mechanism \citep{lambrecht2019algorithmic}, etc.
Even though algorithms could lead to various bias issues, implementing appropriate designs and, regulations can make them positive forces for equity \citep{kleinberg2018discrimination, chouldechova2017fair, rudin2019stop}.

To deal with bias problems in human and machine decision making. researchers have come up with diverse methods. The most direct way is to obtain and include more useful data. \cite{kleinberg2018algorithmic} show direct inclusion of a protected variable (e.g., race) is useful for mitigating unfairness. \cite{lu2019value} find that using proper alternative data could improve both financial profitability and equality. Aside from enriching the data, learning from repeated events can also mitigate bias. \cite{cai2016judging} conclude that based on signaling theory, evaluators/investors will leverage information from repeated borrowing of the same borrow in her lending history. \cite{kim2020importance} argues that borrowers’ past track record within the platform have the most important impact (than other demographic factors) on predicting the repayment performance of their current loans.

Another flourishing track to combat decision bias is to	invent more transparent, delicate and de-biasing algorithms. It has been widely shown that enhanced algorithmic transparency and interpretability can help eliminate bias, and sometimes simple and transparent models are able to outperform complex black-box models \citep{rudin2020broader, rudin2019globally, rudin2019stop, hu2019optimal}. \cite{wang2013bayesian} demonstrate that Bayesian  investment model can significantly improve investors' investment decisions based on other investment models. \cite{choudhury2020machine} argue that human capital and machine learning can complement each other through combining algorithms and domain expertise or knowledge. Many other de-biasing methods draw from the perspective of statistics, optimizations, and behaviors, etc \citep{berk2017convex, lum2016statistical, hardt2016equality, kamiran2010discrimination, fu2021crowds}.

In terms of decision quality, human predictions often tend to be less accurate, which can negatively affect the quality of their decisions. This is because on the one hand, people may have resource-limited brain to process complex computation in evaluation \citep{icard2018bayes}; on the other hand, people may use a simple updating rule, which, for example, linearly combines their personal experience and accumulated knowledge for repeated tasks \citep{jadbabaie2012non}.

Compared with human decision making, algorithm-based decision making has demonstrated superior ability to achieve better accuracy and handle more complex information. Human v.s. machine decision making has been widely studied in healthcare area. In most cases, machine learning models outperformed or tied the judgment accuracy of an average clinician  \citep{camerer201924}, and only a small fraction of clinicians were more accurate than machine learning models \citep[et al.]{goldberg1970man}.
Mechanical-prediction techniques were about 10\% more accurate than clinical predictions \citep{grove2000clinical}. Very simple actuarial method (i.e., linear combination of criterion variables) has been shown to consistently perform  than clinical judgment \citep{dawes1971case}.

However, there are also scenarios in which human experts can outperform machines. Some examples are tasks that heavily require theory-driven judgement that are not suitable for statistical models; rare events or outliers that have never been seen by algorithms; complex configural relationships between the features and the dependent variable \citep{dawes1989clinical}.
In our context, micro-loan approval decisions do not face these problems that make humans better than machines. And in finance industry, algorithms are widely used to identify credit risks, with XGBoost \citep{chen2016xgboost} as the most popular one. Using XGBoost as our focus machine learning model, we examine its bias inherited from human decisions.

Methodologically, our paper also builds upon the abundant work on modeling human decision dynamics through structural models. \cite{erdem2008dynamic} use Bayesian learning framework to model consumers' brand choices under quality signals from advertisement, price and past consumption experiences.
\cite{huang2014crowdsourcing} investigate the dynamic of users' idea posting behavior on a crowd sourcing platform.
\cite{zhang2019structural} study participants' learning behavior from superstars in crowdsourcing contests.
\cite{zhang2020learning} examine taxi drivers' learning behavior based on fine-grained GPS observations.
In this paper, we model the learning dynamics in loan application evaluators' decisions, and disentangle and estimate preference-based bias and belief-based bias in their behavior.

\section{Research Context and Data}

\subsection{Context}

We obtained our data from a leading Asian micro-lending platform. The platform was founded in 2011 and offers microloans at an average size of approximately \$450 USD. Loan applicants on the platform use the loans primarily to fulfill temporary financial needs including supplementary working capital for small businesses, irregular shopping needs, education spending, and medical expenses. To apply for a loan, applicants must provide their personal information such as name, gender, age, income level, and a copy of their national identity cards. They must also provide their contact persons. People under the age of 18 and all students at school or university are not qualified to apply as they usually have no independent income sources. The loan term ranges from one to eight months. The annual interest rate charged by the platform is approximately 18\% (plus or minus 1\%, depending on the credit line of the borrowers).

\begin{figure*}
\centering
    \centering
    \begin{subfigure}[b]{0.49\textwidth}
        \centering
        \includegraphics[trim={0mm 0mm 0mm 0mm}, width=\textwidth]{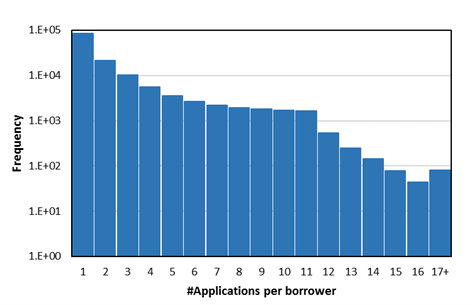}
        \caption[]%
        {{\footnotesize Frequency of loan applications}}
        \label{fig:freq-application}
    \end{subfigure}
    \begin{subfigure}[b]{0.49\textwidth} 
        \centering 
        \includegraphics[trim={0mm 0mm 0mm 0mm}, width=\textwidth]{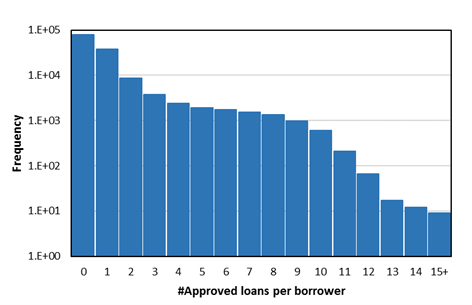}
        \caption[]%
        {{\footnotesize Number of approved loans}}
        \label{fig:freq-approve}
    \end{subfigure}
\caption{Statistics of Loan Applications}
\label{fig:application}
\end{figure*}

During our sample period, the platform evaluates applicants’ credit risk manually by its employees (i.e., evaluators). All the evaluators are trained regularly to maintain consistent evaluation criteria, which are derived from their collective daily work experience. Besides, no gender bias training has been conducted by the platform. They do \textit{not} use any AI technologies (e.g., machine learning) as automatic or auxiliary tools for evaluation. In specific, after an applicant fills in her personal information and submits the application in the system, he will be randomly assigned to an evaluator. Then, the evaluator decides whether to issue the loan by assessing whether the applicant can bring positive economic benefit (i.e., profit) to the platform. Loan profit is roughly  calculated based on the predicted probabilities of delinquency and default. If a borrower fails to repay an installment, he will be regarded as delinquent. If a loan is unpaid 90 days or more after the due date, default is confirmed by the platform. \footnote{These definitions and operations are similar to those in other literature such as \cite{drozd2017modeling} and \cite{pope2011s}.}

The estimations of the chances of delinquency and default are based on the collected personal information for all new applicants. For repeated applicants, evaluators will additionally leverage information from the applicants’ repayment performance on previous loans, such as the final overdue days (denoted as $D$), the proportion of overdue installments (denoted as $M$), the proportion of installments with positive attitude from the borrower (denoted as $A$), which is measured from the records of whether a borrower has presentedshown a positive attitude fortowards their financial obligation during his or her communication with the platform, and the proportion of installments with financial help from family or friends (denoted as $H$).

When a borrower becomes delinquent, the platform will impose financial penalty on them. Simultaneously, debt collection methods such as sending reminder notifications to them and their contact persons will be implemented. Borrowers in default are prohibited from applying for loans again on the focal platform. Default records are also submitted to the personal credit record system maintained by the central government and a shared blacklist system maintained by a symposium of micro-finance institutions. The platform may take legal actions against defaulters.

\subsection{Data and Description}

Our data set contains fine-grained information of both the applicants whose submitted applications were approved and those whose applications were rejected by the platform between January 2015 and September 2017 (i.e., 33 months). During the sample period, there are 311,200 loan applications in total, among which 135,938 loan applications (taking up or approval rate 43.68\%)  were approved, whereas 175,263 were rejected by the platform. Our sample covers 139,454 borrowers; that is, the average number (frequency) of loan applications per borrower is 2.23 (= 311,200/139,454). In our sample, 53,503 (38.37\%) borrowers applied more than once, and they contributed 225,248 applications in total (i.e., 4.21 on average per borrower). For these multiple-time borrowers, the average approval rate is 47.24\%. The average approval rate is only 34.34\% for the 85,951 borrowers who applied just once. This indicates the platform’s preference towards repeated borrowers, which is reasonable as these borrowers have performed well in historical loans. Figures~\ref{fig:freq-application} and \ref{fig:freq-approve} display the distributions of the frequency of loan applications and number of approved loans respectively.

For all the applicants, we obtain their demographic and socioeconomic data including gender, education level, monthly income level, disposable personal income per capita (DPI) of their home city, and house ownership, and their loan information including loan amount, loan term, and annual interest rate. We likewise have detailed per-installment repayment information of the approved loans. 
Table~\ref{tab:application} summaries information of the approved and rejected applications.

\begin{table*}
\begin{tabular}{llllll}
\hline
\multirow{2}{*}{}              & \multirow{2}{*}{Observations} & \multicolumn{2}{l}{Repeated applicants} & \multicolumn{2}{l}{New applicants} \\ \cline{3-6} 
                               &                               & Mean                & S. D.             & Mean             & S. D.           \\ \hline
Gender (1 = female, 0 = male)  & 53,503/85,952                 & 0.19                & 0.39              & 0.18             & 0.39            \\
Education level                & 53,503/85,952                 & 2.25                & 0.64              & 2.27             & 0.63            \\
Monthly income level           & 53,503/85,952                 & 3.27                & 1.80              & 3.37             & 1.90            \\
Home city DPI                  & 53,503/85,952                 & 2.34                & 1.31              & 2.54             & 1.49            \\
House ownership (1 = self-own) & 53,503/85,952                 & 0.17                & 0.38              & 0.16             & 0.37            \\
Loan amount (USD)              & 311,200                       & 460.70              & 81.59             & 458.78           & 321.30          \\
Loan term (month)              & 311,200                       & 5.88                & 1.54              & 5.86             & 1.35            \\
Yearly interest rate (\%)      & 311,200                       & 14.05               & 1.28              & 14.36            & 1.44            \\ \hline
\multicolumn{6}{l} {\footnotesize Note: Education level: 1 = middle school; 2 = technical school; 3 = undergraduate; 4 = postgraduate.} 
\end{tabular}
\caption{Information of the Approved and Rejected Applications}
\label{tab:application}
\end{table*}

\subsection{Model-Free Analyses}

\begin{figure*}
\centering
    \centering
    \begin{subfigure}[b]{0.32\textwidth}
        \centering
        \includegraphics[trim={0mm 0mm 0mm 0mm}, width=\textwidth]{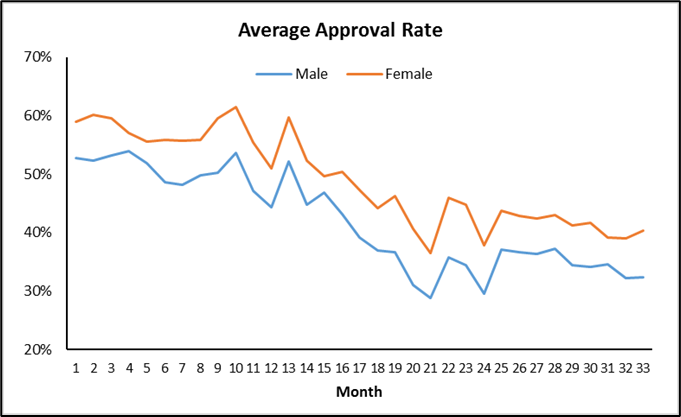}
        \caption[]%
        {{\scriptsize Average approval rate for all applicants
        }}
        \label{fig:time-approval}
    \end{subfigure}
    \begin{subfigure}[b]{0.33\textwidth} 
        \centering 
        \includegraphics[trim={0mm 0mm 0mm 0mm}, width=\textwidth]{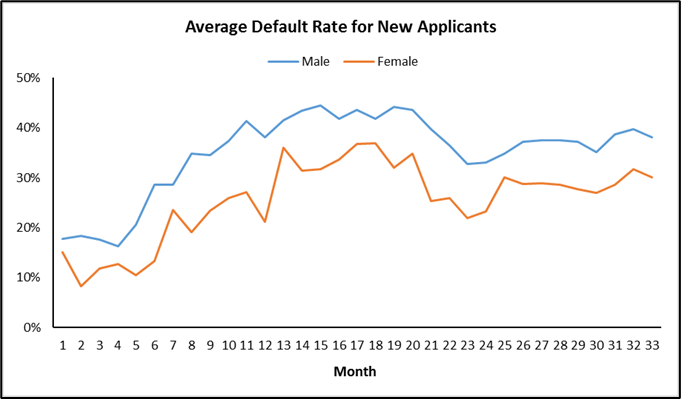}
        \caption[]%
        {{\scriptsize Average default rate for new applicants
        }}
        \label{fig:time-default-new}
    \end{subfigure}
     \begin{subfigure}[b]{0.33\textwidth} 
        \centering 
        \includegraphics[trim={0mm 0mm 0mm 0mm}, width=\textwidth]{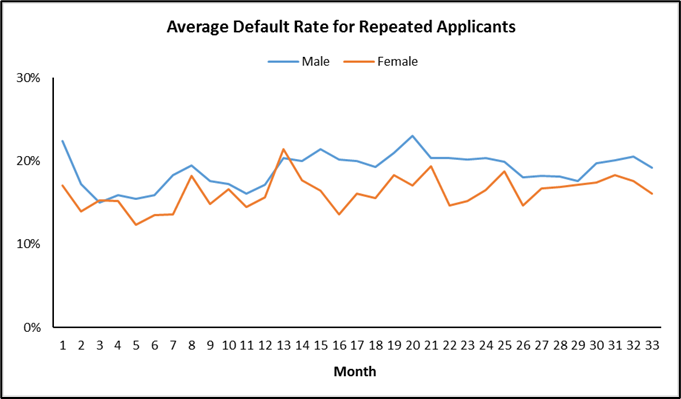}
        \caption[]%
        {{\scriptsize Average default rate for repeated applicants
        }}
        \label{fig:time-default-repeat}
    \end{subfigure}
\caption{Time Trends of Approval and Default Rate}
\label{fig:approval-default}
\end{figure*}

On the platform, female applicants are generally more likely to be approved. The gaps between female and male approval rates do not shrink over time(Figure~\ref{fig:time-approval}), implying that the platform evaluators may have a persistent impression of credit risks between males and females.

\begin{figure}
\centering
        \centering
        \includegraphics[trim={0mm 0mm 0mm 0mm}, width=0.5\textwidth]{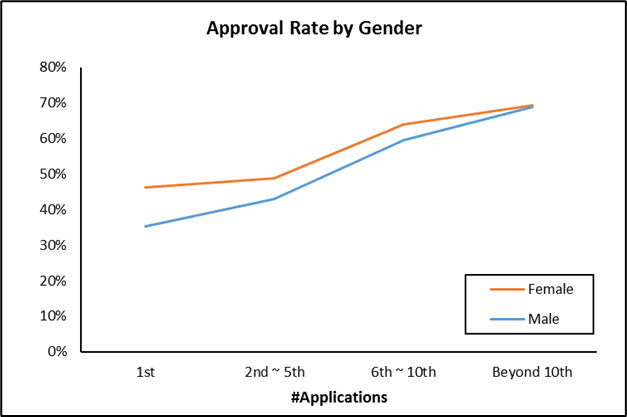}
\caption{Approval Rate by Gender}
\label{fig:application}
\end{figure}

\begin{figure*}
\centering
    \centering
    \begin{subfigure}[b]{0.49\textwidth}
        \centering
        \includegraphics[trim={0mm 0mm 0mm 0mm}, width=\textwidth]{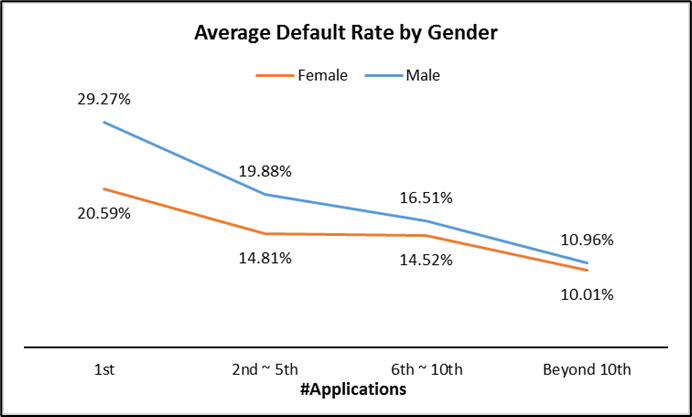}
        \caption[]%
        {{\footnotesize Average Default Rate by Gender for users with different number of applications }}
        \label{fig:default-application}
    \end{subfigure}
    \begin{subfigure}[b]{0.49\textwidth} 
        \centering 
        \includegraphics[trim={0mm 0mm 0mm 0mm}, width=\textwidth]{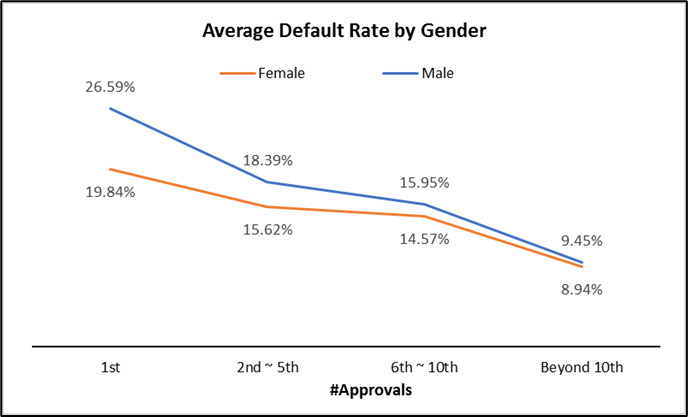}
        \caption[]%
        {{\footnotesize Average Default Rate by Gender for users with different number of approvals}}
        \label{fig:default-approval}
    \end{subfigure}
\caption{Average Default Rate by Gender}
\label{fig:default}
\end{figure*}

When we consider only borrowers who have applied repeatedly (Figure~\ref{fig:application}), the approval gender gap shrinks, suggesting that learning helps amend the platform evaluators’ prior bias in gender. 
Figure~\ref{fig:default} shows the trend of the default rate as borrowers as the number of applications or the number of their previously approved loans increases. Consistently, the gender gap in the default rate is smaller for repeated borrowers; and it keeps decreasing as the borrowers’ number of previous applications increases, and as number of previous approvals increases. This is consistent with the decreasing gender gap in the approval rate, which indicates that evaluators can learn effectively from users’ previous repayment behaviors and adjust their prior belief-based bias in gender.

\begin{figure}[]
\centering
    \centering
    \begin{subfigure}[b]{0.4\textwidth}
        \centering
        \includegraphics[trim={0mm 0mm 0mm 0mm}, width=\textwidth]{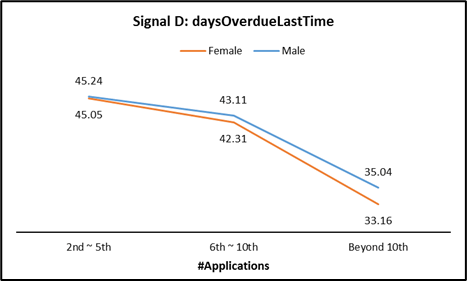}
        \caption[]%
        {{\footnotesize Signal D by Gender for users with different number of applications }}
    \end{subfigure}
    \begin{subfigure}[b]{0.4\textwidth}
        \centering
        \includegraphics[trim={0mm 0mm 0mm 0mm}, width=\textwidth]{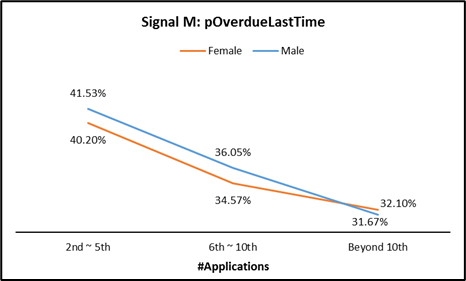}
        \caption[]%
        {{\footnotesize Signal M by Gender for users with different number of applications }}
    \end{subfigure}\\
    \begin{subfigure}[b]{0.4\textwidth}
        \centering
        \includegraphics[trim={0mm 0mm 0mm 0mm}, width=\textwidth]{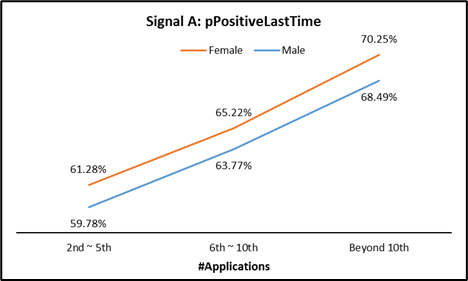}
        \caption[]%
        {{\footnotesize Signal A by Gender for users with different number of applications }}
    \end{subfigure}
    \begin{subfigure}[b]{0.4\textwidth}
        \centering
        \includegraphics[trim={0mm 0mm 0mm 0mm}, width=\textwidth]{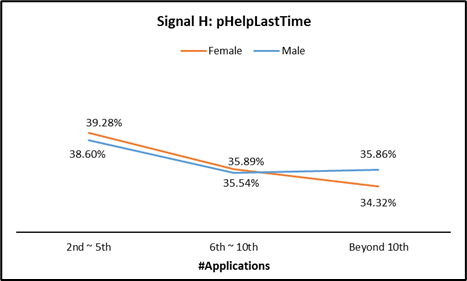}
        \caption[]%
        {{\footnotesize Signal H by Gender for users with different number of applications }}
    \end{subfigure}
\caption{Signals Across Number of Applications}
\label{fig:signal}
\end{figure}

As noted earlier, the evaluator relies on four signals from the users' past repayment behaviors to make approval decisions on loan applications. Figure~\ref{fig:signal} shows the values of the four signals against the number of previous applications by genders. Generally, we find that, given the number of previous applications, all the four signals show similar values between females and males, indicating that these signals may help the platform evaluators to adjust their prior bias in gender.

\section{Model}
\label{sec:model}

We consider a model in which the loan platform decides whether or not to approve a loan application by applicant $i$ at time $t$ (here $t$ indicates the $t$-th application rather than a natural time unit). We model the evaluator behavior in a dynamic environment where the evaluator is uncertain about the true credit quality of applicants. 
When a new borrower comes to the micro-lending platform, the evaluator only observes her demographics and form a prior belief based on the demographic data. For every borrower, without any previous repayment behavior being observed, her first application is preprocessed by the evaluator using the prior belief. If a borrower's loan application is approved at $t$, then at the time of her next loan application, i.e., $t+1$, the evaluator will use the previous repayment behaviors as additional signals of the borrower's credit quality. 

At time $t=0$, without any observation of borrower $i$'s repayment behaviors,
given $i$'s demographics $\mathbf{X}_i$, 
the evaluator forms a belief of her credit quality with mean ${\beta} \mathbf{X}_i$ and variance $\sigma_{Q_0}^2$. $\mathbf{X}_i$ includes 
\textit{gender, age, education level, marriage status, number of children, house ownership, monthly income, the disposable personal income (DPI) of borrowers’ living cities (in 2017)}. We also incorporate a constant 1 and \textit{the time of $i$'s first application} into $\mathbf{X}_i$, because the overall borrowers' qualities may keep changing over time. We assume that the prior belief follows a normal distribution:

\begin{equation}
    Q_{i0} \sim N({\color{black} \bm{\beta} \mathbf{X}_i,\sigma_{Q_0}^2}) \text{  for }i=1,...,N.
\end{equation}

\noindent where $\bm{\beta} \mathbf{X}_i$ is the evaluator's subject prior belief about the mean credit quality of a borrower with demographics $\mathbf{X}_i$. The coefficient for gender is ${\beta}_{g}$, where gender $g \in \{M, F\}$, $M$ stands for males and $F$ stands for females. We normalize ${\beta}_{F}$ to be zero. Therefore ${\beta}_{M}$ captures the \textbf{belief-based bias}.
For every borrower, we assume that the evaluator believes that their credit qualities are of identical uncertainty, i.e., $\sigma_{Q_0}^2$ is identical for all $i$.

When the evaluator processes the loan application of borrower $i$ at time $t$, she has an information set $I_{it}$, which contains all loan repayment histories up to the loan of time $t-1$. Given the information set, the evaluator form an posterior belief of $Q_{i}$. Let $Q_{it}$ be the expectation of the evaluator's belief for $i$'s credit quality at time $t$, i.e.

\begin{equation}
    Q_{it} = \mathrm{E}[Q_{i}|I_{it}]
\end{equation}

At time t, when the evaluator is reviewing the loan application from borrower $i$, she observes four signals from $i$'s repayment behavior on previous loans.
Consider when an applicant $i$ applies for a loan at time $t$, the evaluator observes four signals of the repayment behavior of her last loan, the final overdue days $D_{it}$, the proportion of overdue installments $M_{it}$, the proportion of installments with positive attitude from the borrower $A_{it}$, and the proportion of installments with financial help from family or friends $H_{it}$. Some of these signals may be more informative than others. Therefore, we tried several different combinations of these signals, and drop $M_{it}$ from our signal list in our final model as $M_{it}$ turns out to be uninformative and does not affect evaluators’ loan approval decisions significantly..

In previous literature \citep[et al.]{erdem2008dynamic, huang2014crowdsourcing, zhang2020learning}, it is common to model such behaviors using Bayesian updates of a normal-normal conjugate prior-posterior. We also start with such a conventional normal-normal conjugate Bayesian model, i.e., each time when additional signals become available, the evaluator updates her belief of a borrower’s credit quality in a Bayesian fashion. However, the estimated parameters of such a Bayesian learning model show that all variances of these signals are extremely small. This indicates that evaluators weight the most recent signals heavily and are not updating their beliefs in a Bayesian fashion. Therefore, we use a simplified model to capture the evaluators' updating behaviors, where the posterior belief is a weighed sum of the prior belief and the new signals. We tried different combinations of the four signals. 
The weighted sum model with signals $D_{it}$, $A_{it}$, and $H_{it}$ performs the best and achieves the highest likelihood. We use this mode as our main model. In this section, we introduce this model in detail; and in the appendix (Section~\ref{sec:appendix}), we summarize the detailed of the Bayesian one. 

Because the final overdue days $D_{it}$ has a long-tail distribution, we take the logarithm of $D_{it}$ and use $\log D_{it}$ for subsequent calculations. We assume the evaluators believe that all these signals are linearly related to the credit quality in the following way:

\begin{equation}
\begin{aligned}
    \log D_i^M &= D_0 + \phi Q_i, \\
    A_i^M &= A_0 + \psi Q_i, \\
    H_i^M &= H_0 + \rho Q_i,
\label{eq:signal-mean-quality}
\end{aligned}
\end{equation}

\noindent where $\phi$, $\psi$, $\rho$ are parameters of slopes, $D_0$, $A_0$, and $H_0$ are corresponding intercepts, and $\log D_i^M$, $A_i^M$, and $H_i^M$ are the means of $D_{it}$, $A_{it}$, $H_{it}$. We assume each borrower's signals are distributed surrounding their means.
At each time $t$ when the evaluator observes these repayment behaviors, she updates her belief about the credit quality based on a weighted sum of the prior and the signals:

\begin{equation}
\begin{aligned}
    Q_{it} = &(1 - \alpha_D -\alpha_A - \alpha_H) Q_{i,t-1} + \\
    & \alpha_D \frac{\log D_{it} - D_0}{\phi} + \alpha_A \frac{A_{it} - A_0}{\psi} + \alpha_H \frac{H_{it} - H_0}{\rho}
\label{eq:quality-weighted-sum}
\end{aligned}
\end{equation}

\noindent where $\alpha_D, \alpha_A, and \alpha_H$ are the wights assigned to the three signals. 
At time $t$, the evaluator decides whether to approve $i$'s loan application based on her updated belief of the applicant's quality $Q_{it}$. The evaluator first calculates the probability of non-default through a sigmoid function:

\begin{equation}
\begin{aligned}
    p_{it} = h(Q_{it}) = \frac{1}{1+\exp(-Q_{it})}.
\end{aligned}
\end{equation}

Then, with the probability of non-default $p_{it}$, the evaluator decides to approve or reject $i$'s loan application of time $t$ through a utility function. There are two key components in the utility function (Equation~\ref{eq:utility}). The first component captures the expected profit of approving this loan. The second component contains the evaluator's preference based bias. And we also assume there is a random shock within the utility function. The utility function is as follows:

\begin{equation}
\begin{aligned}
    u_{it} = z*(p_{it}a_{it}-(1-p_{it})b_{it})-c_{ig}+\epsilon_{it},
\end{aligned}
\label{eq:utility}
\end{equation}

\noindent where $g \in \{M, F\}$, and $M$ stands for males and $F$ stands for females, $c_{ig}$ is the preference-based bias with $c_{iF}$ normalized to be zero. Therefore, $c_{iM}$ captures the \textbf{preference-based bias}, which persists and is not affected by observing new signals. $a_{it}$ is the profit earned by the platform if the loan is paid back, and $b_{it}$ is the loss the platform incurs if the loan defaults. Both $a_{it}$ and $b_{it}$ are observed values in our dataset. $z$ is the price parameter (or marginal utility of money). $p_{it}$ is the non-default probability of $i$ at time $t$, which is related to its quality $Q_{it}$. All these parameters are estimated through maximizing the likelihood.

\subsection{Estimation Results}

We report parameter estimates for our model in Table~\ref{tab:estimates}. Both the preference-based bias $c_{iM}$ (we normalize $c_{iF}$ to be zero) and the belief-based bias ${\beta}_{M}$ have expected signs ($c_{iM}=0.2309, {\beta}_{M}=-0.1042$), implying that the evaluator has a preference for female loan applicants. The estimate of the belief based bias ${\beta}_{M}$ is significantly negative, suggesting that evaluators have a lower prior belief for males' credit qualities. The estimate of $c_{iM}$ is significantly positive. This implies  that there is a significant preference-based bias in gender that favors female applicants, which cannot be corrected by observing repayment behaviors.

Apart from ${\beta}_{F}$, all other $\beta$s also have expected signs. This is consistent with the evaluator's preference for applicants with a better socioeconomic status as we observe in the data.

\begin{table}[]
\centering
\begin{tabular}{cll}
\hline
\multicolumn{1}{l}{\textbf{Paramteters}} & \textbf{Estimate} & \textbf{Std. error} \\ \hline
\multicolumn{1}{l}{$c_{iM}$}             & 0.2390$^{***}$  &  0.0220             \\ \hline
\multicolumn{3}{l}{Signal D}                                                       \\
$\phi$                                      &-0.0138$^{***}$  &  0.0008              \\
$D_0$                                       & 0.5219$^{***}$  &  0.0207              \\ \hline
\multicolumn{3}{l}{Signal A}                                                       \\
$\psi$                                      & 0.3492$^{***}$  &  0.0034              \\
$A_0$                                       & 0.3788$^{***}$  &  0.0063              \\ \hline
\multicolumn{3}{l}{Signal H}                                                       \\
$\rho$                                   & 0.9803$^{***}$  &  0.0800             \\
$H_0$                                       & 0.1252$^{***}$  &  0.0399              \\ \hline
\multicolumn{1}{l}{z}                    & 0.0154$^{***}$  &  0.0001     \\ \hline
\multicolumn{3}{l}{Coefficients of the prior $\bm{\beta}$}                                                   \\
${\beta}_0$                 & -0.8961$^{***}$  &  0.0119              \\
${\beta}_{M}$         & -0.1042$^{***}$  &  0.0088              \\
${\beta}_{firstAppMonth}$                            & -0.0201$^{***}$  &  0.0003              \\
${\beta}_{housing}$                                  & 0.1458$^{***}$  &  0.0058              \\
${\beta}_{education}$                                & 0.2443$^{***}$  &  0.0038              \\
${\beta}_{income}$                                   & 0.0936$^{***}$  &  0.0013              \\
${\beta}_{DPI}$                                      & 0.1176$^{***}$  &  0.0018              \\ \hline
\multicolumn{1}{l}{$\alpha_D$}           & 0.0097$^{***}$  &  0.0007          \\
\multicolumn{1}{l}{$\alpha_A$}           & 0.9780$^{***}$  &  0.0058              \\
\multicolumn{1}{l}{$\alpha_H$}           & 0.0121$^{***}$  &  0.0009              \\ \hline
\multicolumn{3}{l} {\footnotesize Note: *$p < 0.1$; **$p<0.05$; ***$p<0.01$}
\end{tabular}\\
\caption{Structural Model Estimation Results}
\label{tab:estimates}
\end{table}

Our estimates of the slopes in the signal D-quality, signal A-quality and signal H-quality relationships are negative ($\phi=-0.0138$), positive ($\psi=0.3492$) and positive ($\rho=0.9803$) respectively. These results suggest that in the evaluator's decision-making process, larger final overdue days $D_{it}$ are associated with poor credit quality; while the proportion of installments with a positive attitude $A_{it}$ is positively associated with loan approvals. Getting financial help from family and friends $H_{it}$ is also viewed by evaluators as a positive signal for credit quality and is associated with a lower default probability.

As can be seen in Equation~\ref{eq:quality-weighted-sum}, the evaluator updates her belief based on a weighted sum of the prior belief and the signals. The estimates of the weights $\alpha_D=0.0097$, $\alpha_A=0.9780$ and $\alpha_H=0.0121$ indicate that the evaluator gives most weight to the signal $A_{it}$, with a wight as high as 0.9780.

In Table~\ref{tab:SM-actual}, we compare the characteristics of the actual observed approved users and the expected values of the characteristics of the approved users our structural model predicts. All these statistics are very similar between the actual observations and our model’s predictions. This suggests our model captures the decision process well.

\begin{table*}
\centering
\begin{tabular}{ccccccccccccc}
\hline
 & \multicolumn{2}{c}{number of users} & \multicolumn{2}{c}{number of females} & \multicolumn{2}{c}{mean of housing} & \multicolumn{2}{c}{mean of DPI} & \multicolumn{2}{c}{mean of education} & \multicolumn{2}{c}{mean of income} \\ \cline{2-13} 
t               & SM                & actual          & SM                 & actual           & SM               & actual           & SM             & actual         & SM                & actual            & SM               & actual          \\ \hline
1                  & 53539.33          & 51019           & 12113.01           & 11797            & 0.085023         & 0.082192         & 1.078531       & 1.038077       & 0.967054          & 0.926234          & 1.525907         & 1.470026        \\
2                  & 22371.7           & 22486           & 4742.72            & 4690             & 0.03179          & 0.031874         & 0.462848       & 0.464454       & 0.389653          & 0.391233          & 0.621982         & 0.626156        \\
3                  & 14110.73          & 13993           & 2802.512           & 2734             & 0.018555         & 0.018551         & 0.277881       & 0.27404        & 0.240154          & 0.237777          & 0.379792         & 0.376518        \\
4                  & 10139.22          & 10392           & 1967.239           & 2015             & 0.013            & 0.013295         & 0.192308       & 0.194846       & 0.17002           & 0.174488          & 0.264952         & 0.270864        \\
5                  & 7917.322          & 8335            & 1520.431           & 1605             & 0.009977         & 0.010598         & 0.146994       & 0.153377       & 0.1317            & 0.138655          & 0.202175         & 0.211797        \\
6                  & 6376.219          & 6915            & 1222.235           & 1314             & 0.008252         & 0.009007         & 0.117074       & 0.125568       & 0.105735          & 0.114719          & 0.161333         & 0.174509        \\
7                  & 5035.121          & 5481            & 970.1422           & 1056             & 0.006301         & 0.006884         & 0.092232       & 0.099043       & 0.083498          & 0.090826          & 0.126435         & 0.137056        \\
8                  & 3629.12           & 3968            & 692.8316           & 755              & 0.004384         & 0.004905         & 0.065735       & 0.0714         & 0.060267          & 0.065685          & 0.090112         & 0.098269        \\
9                  & 2342.886          & 2564            & 441.4362           & 484              & 0.002798         & 0.003091         & 0.042753       & 0.046345       & 0.038917          & 0.042602          & 0.057552         & 0.062881        \\
10                 & 1150.928          & 1272            & 219.6785           & 234              & 0.001276         & 0.001413         & 0.020986       & 0.02309        & 0.019207          & 0.021233          & 0.029364         & 0.032297        \\
11                 & 436.4425          & 475             & 80.60686           & 89               & 0.000486         & 0.000552         & 0.00845        & 0.009093       & 0.007253          & 0.007866          & 0.011423         & 0.012348        \\
12                 & 130.8604          & 129             & 21.06548           & 20               & 0.000149         & 0.000136         & 0.002757       & 0.002718       & 0.002312          & 0.002273          & 0.003748         & 0.003571        \\
13                 & 47.87013          & 51              & 8.762629           & 9                & 3.74E-05         & 3.59E-05         & 0.000959       & 0.001011       & 0.000824          & 0.000875          & 0.001393         & 0.001492        \\
14                 & 14.98715          & 15              & 1.997452           & 2                & 1.35E-05         & 1.43E-05         & 0.00036        & 0.000359       & 0.000251          & 0.000251          & 0.000443         & 0.000437        \\
15                 & 5.211481          & 5               & 0                  & 0                & 0                & 0                & 0.00013        & 0.0001         & 9.28E-05          & 9.32E-05          & 0.000181         & 0.000172        \\
16                 & 0.999583          & 1               & 0                  & 0                & 0                & 0                & 5.73E-05       & 5.74E-05       & 1.43E-05          & 1.43E-05          & 2.87E-05         & 2.87E-05        \\ \hline
\end{tabular}
\caption{Comparison of Simulated and Actual Characteristics of Approved Users}
\label{tab:SM-actual}
\end{table*}

\section{Policy Simulations}
\label{sec:policy}
We conduct several sets of counterfactual simulations to evaluate the effects of eliminating the biases found in the data on the outcome of loan applications across different gender groups. Our counterfactual analyses are done on a second dataset. It covers all the loan records from a one-month experimental period (“full sample” hereafter). During this period, all applicants are approved without screening. As a result, we have true label of all users. This ensures our results are based on the entire user distribution, rather than just the approved users, which have a different distribution from the whole user pool.

Specifically, we calculate the expected profits of the platform based on the predicted approval probability by a number of variants of the estimated structural model. We also examine the gender gap in the approval true positive rate (TPR). We adopt the concept of \textit{equal opportunity}, one of the most popular fairness notions, to measure the extent of bias. \textit{Equal opportunity} requires that qualified individuals, no matter what their sensitive attributes are, have an equal opportunity to receive favorable outcomes. In our loan application setting, this means two gender group should have the same true positive rates. We find that the elimination of either the preference-based bias or the belief-based bias can simultaneously increase the platform’s profit and reduce the gender gap in loan approval decisions (TPR).
In Section~\ref{sec:ml-bias}, We also feed the counterfactual datasets into machine learning algorithms to see how machine captures these different behaviors.

\begin{figure*}
\centering
        \centering
        \includegraphics[trim={10mm 20mm 10mm 20mm}, width=0.6\textwidth]{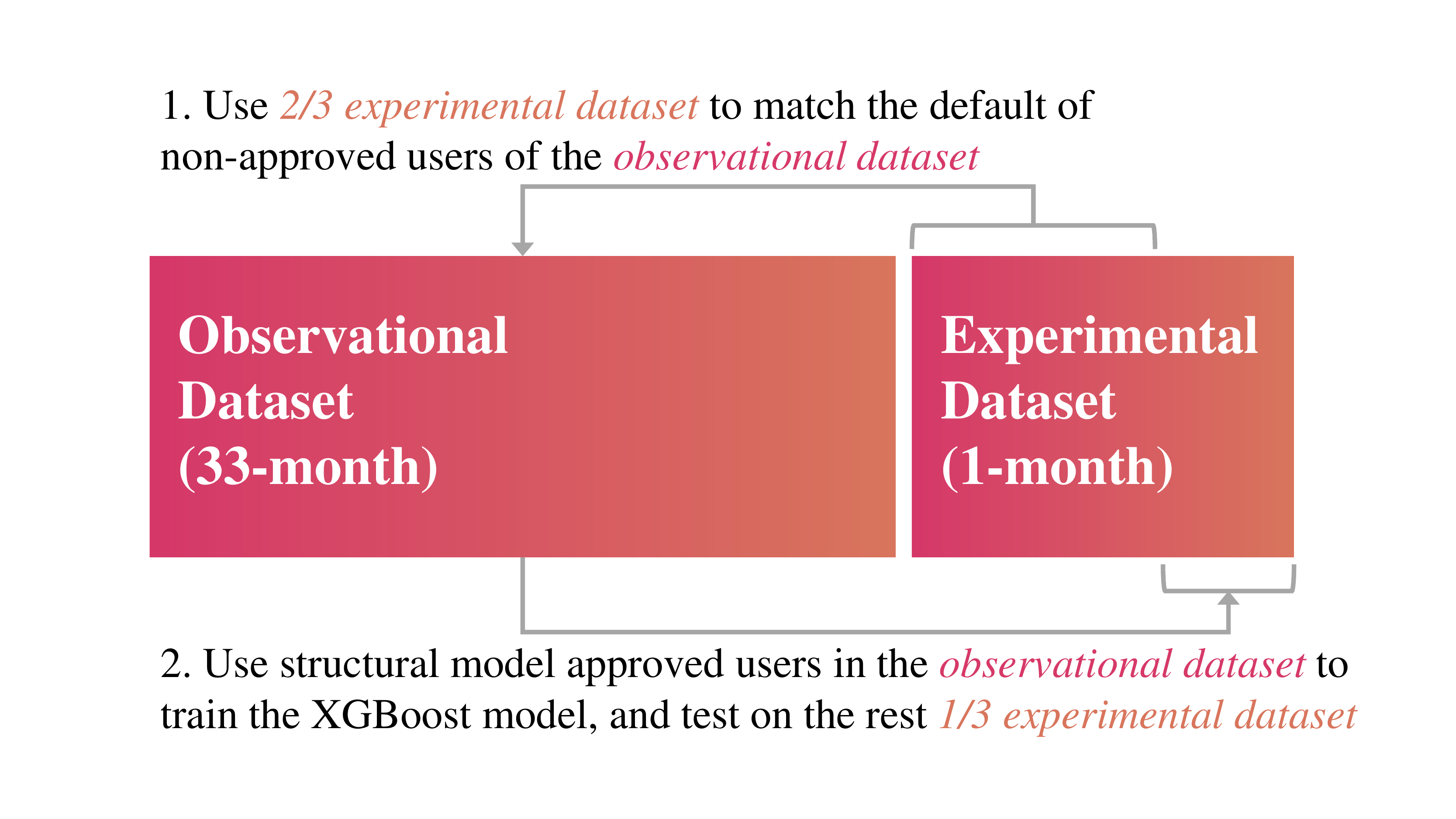}
\caption{Two datasets used in the training and evaluation}
\label{fig:dataset}
\end{figure*}

%
%
%
%
%
%

\subsection{Does eliminating preference-based bias help with the decision making? How does it affect the platform’s payoff?}

As noted above, the preference-based bias refers to people's animus towards a particular group. No matter in reality how well this certain group behaves, people with preference-based bias always make judgements and decisions by prejudice. In our setting, we focus on the preference-based bias in gender. In our model, $c_{ig}$ captures the preference-based bias. For applicant $i$ of gender $g \in \{M, F\}$ ($M$  for males and $F$ for females), we normalize $c_{iF}$ for females to be zero. And the estimation result for preference-based bias is $c_{iM}=0.2390$. Note that in Equation~\ref{eq:utility}, we have the profit minus sign in front of $c_{ig}$, therefore $c_{iM}=0.2390$ suggests the evaluator has a preference for female borrowers and a prejudice against male borrowers.

In an ideal setting, all the evaluators are trained and the prejudice to gender is completely removed. The elimination of the preference-based bias can be operationalized by setting $c_{iM}$ to be zero. We simulate the evaluators' decisions with $c_{iM}=0$ but all other parameters unchanged. We then compare the true positive rate (TPR) and the payoff under the original decision process and the one with preference-based bias removed on our full sample.

When the evaluators make approval decisions without the preference-based bias in gender, the TPR for male users are all higher than their corresponding value from the original decision process. 
Note that for female users, $c_{iF}=0$, thus their TPR stays the same under two different decision-making process. In sum, eliminating the preference-based bias can generally decrease the TPR gap between the two gender groups in our setting. But this decrease is relatively small (from 9.69\% to 6.94\%, Table~\ref{tab:counterfactual-gap}).




We also compare the platform's expected welfare (profit) under the current decision-making process and the one with preference-based bias removed. We observe that by eliminating the preference-based bias ($c_{iM}$) in the loan approval process, the platform obtains a higher profit (Table~\ref{tab:counterfactual-profit-all}). The increase in the profit results from better decisions made on male applicants. Specifically, the increase driven by the gain from lowering the approval probability for male borrowers who eventually default on loans, which exceed the loss of lowering the approval probability for nondefault male borrowers.
These findings suggest that although the current decision-making process incorporates gender information to identify high quality users, the evaluators over underrate male users and disdain male borrowers too much. Therefore, the preference-based bias results in suboptimal decisions.


\subsection{Does removing gender from the prior belief formation help with the decision making? How does it affect the platform’s payoff?}

In this subsection, we examine the effects of removing the gender information in the prior belief, i.e. set $\beta_M$ to be zero. $\beta_{g}, g \in \{M, F\}$ ($M$  for males and $F$ for females) captures the belief-based bias. 
Belief-based bias refers to evaluators’ subjective beliefs about certain groups. This kind of belief can be updated through observing additional behavioral signals, i.e. repayment behaviors in this paper.
In our setting, we normalize female users' $\beta_{F}$ to be zero, therefore $\beta_M$ captures the relative belief-based bias for females compared with males. The estimated value of $\beta_M$ is $-0.1042$, which indicates the evaluators have a subjective prior belief in favor of female borrowers. We compare the TPR and the platform's payoff under the original decision process and the one with belief-based bias removed on our full sample.

Under the current decision making process, we observe a higher TPR for females than males. Since we normalized $\beta_{F}$ to be zero, the TPR of females does not change between the two decision processes. Gender information in the prior decreases the male TPR. With access to the gender information, the evaluators form a prior belief that underrates female borrowers. These additional rejected males are generally of good enough credit qualities. 

\begin{table}[H]
\centering
\begin{tabular}{lll}
\hline
           & $\beta_M=-0.1042$                      & $\beta_M=0$ \\ \hline
$c_{iM}=0.2390$  & 151046.8 & 156957.8  \\ \hline
$c_{iM}=0$ & 155036.7 & 160190.4 \\ \hline
\end{tabular}
\vskip 3mm
\caption{The expected profits of different decision process.}
\label{tab:counterfactual-profit-all}
\end{table}

\begin{table}[H]
\centering
\begin{tabular}{lll}
\hline
           & $\beta_M=-0.1042$                      & $\beta_M=0$ \\ \hline
$c_{iM}=0.2390$  & {9.69\%} & 5.14\%  \\ \hline
$c_{iM}=0$ & {6.94\%} & 2.48\% \\ \hline
\end{tabular}
\vskip 3mm
\caption{The gender gap in the credit evaluation TPR (female's TPR minus male's TPR).}
\label{tab:counterfactual-gap}
\end{table}

\begin{table}[H]
\centering
\begin{tabular}{lll}
\hline
           & $\beta_M=-0.1042$                      & $\beta_M=0$ \\ \hline
$c_{iM}=0.2390$  & {12.31\%} & 5.51\%  \\ \hline
$c_{iM}=0$ & {8.62\%} & 2.04\% \\ \hline
\end{tabular}
\vskip 3mm
\caption{The gender gap in the credit evaluation TPR (female's TPR minus male's TPR). New users only.}
\label{tab:counterfactual-gap-new}
\end{table}

\begin{table}[H]
\centering
\begin{tabular}{lll}
\hline
           & $\beta_M=-0.1042$                      & $\beta_M=0$ \\ \hline
$c_{iM}=0.2390$  & {5.13\%} & 4.10\%  \\ \hline
$c_{iM}=0$ & {3.85\%} & 2.70\% \\ \hline
\end{tabular}
\vskip 3mm
\caption{The gender gap in the credit evaluation TPR (female's TPR minus male's TPR). Repeated users only.}
\label{tab:counterfactual-gap-repeated}
\end{table}


We further investigate the effect of $\beta_{M}$ on the expected profit of the platform. With the belief-based bias removed, the platform obtains a larger profit. This suggests that setting $\beta_{M}$ to zero leads to an increase in the probability of approval for all males, and the gain from lowering the approval probability for default male borrowers dominates the loss from lowering the approval probability for nondefault male users.

On the borrower side, when the belief-based bias is removed, the gender gap in TPR becomes smaller  (Table~\ref{tab:counterfactual-gap}). It decreases from 9.69\% to 5.14\%. This decrease is larger than the one resulting from eliminating the preference-based bias. When the both biases are ruled out, we can achieve the smallest gender gap in the TPR (2.48\%).



\subsection{Are these effects different for new users and repeated users?}

To further investigate the underlying mechanism, we look into the new users and the repeated users separately. Table~\ref{tab:counterfactual-gap-new} shows the TPR gaps in the original and counterfactual settings of the new users, and Table~\ref{tab:counterfactual-gap-repeated} shows the TPR gaps of the repeated users. Consistent with previous observation, eliminating either bias can help mitigate the bias for both new users and repeated users.

We observe that overall the repeated users have smaller bias than new users. This is because of the different characteristic distributions between original new and repeated applicants, and the signals used for learning repeated applicants' credit quality. When removing the preference-based bias but keeping the belief-based bias, human evaluators’ decision bias toward gender decrease from 12.31\% to 8.62\% for new applicants and from 5.13\% to 3.85\% for repeated applicants. If we keep preference-based bias but remove belief-based bias, compared with the former case, human evaluators’ decision bias toward gender has a larger decrease to 5.51\% for new applicants, but a smaller decrease to 4.10\%. This is because for repeated users, human make use of signals to update the prior belief. The belief-based bias becomes less influential for repeated users.

\section{Machine Learning Bias Inherited from Human Bias }
\label{sec:ml-bias}

In this section, we examine how ML algorithms inherit human bias, by feeding the real-word dataset and the counterfactual datasets into ML algorithms. Specially, for the counterfactual setting, we run our structural model with counterfactual parameters to simulate the approval decision; we use the approved users' loan records as the input for our machine learning model. For users who were not approved in reality, we do matching to fill their loan repayment behavior. Here we use XGBoost \citep{chen2016xgboost} as our machine learning model. XGBoost is widely used in loan evaluations and related literature \citep{lu2019value, fu2021crowds}.
As in Section~\ref{sec:policy}, we report the the expected profits (Table~\ref{tab:xgb-profit-all}) and the gender gap of the TPR (Table~\ref{tab:xgb-gap}).

For the XGBoost model, when the preference-based bias is removed, the expected profit increases from 167260.2 to 172125.8 (increases by 4865.6, compared with the structural model's corresponding increase of 3,989.9), and the TPR gap decreases from 7.49\% to 6.20\% (decreases by 1.29\%, compared with the structural model's decrease of 2.75\%). This indicates that machine learning models can mitigate the effects of the preference-based bias.
When the belief-based bias is removed, the expected profit increases from 167260.2 to 171352.0 (by 4091.8, compared with the structural model's increase of 5911), and the TPR gap decreases from 7.49\% to 7.43\% (decreases by 0.06\%, compared with the structural model's decrease of 4.55\%). This indicates that machine learning models can mitigate the effects of the belief-based bias.

\begin{table}[H]
\centering
\begin{tabular}{lll}
\hline
           & $\beta_M=-0.1042$                      & $\beta_M=0$ \\ \hline
$c_{iM}=0.2390$  & 167260.2 & 171352.0 \\ \hline
$c_{iM}=0$ & 172125.8 & 174534.7 \\ \hline
\end{tabular}
\vskip 3mm
\caption{The expected profits of different decision process.}
\label{tab:xgb-profit-all}
\end{table}

\begin{table}[H]
\centering
\begin{tabular}{lll}
\hline
           & $\beta_M=-0.1042$                      & $\beta_M=0$ \\ \hline
$c_{iM}=0.2390$  & {7.49\%} & 7.43\% \\ \hline
$c_{iM}=0$ & {6.20\%} & 5.57\% \\ \hline
\end{tabular}
\vskip 3mm
\caption{The gender gap in the credit evaluation TPR (female's TPR minus male's TPR).}
\label{tab:xgb-gap}
\end{table}

\begin{table}[H]
\centering
\begin{tabular}{lll}
\hline
           & $\beta_M=-0.1042$                      & $\beta_M=0$ \\ \hline
$c_{iM}=0.2390$  & {6.86\%} & 6.63\% \\ \hline
$c_{iM}=0$ & {4.66\%} & 3.98\% \\ \hline
\end{tabular}
\vskip 3mm
\caption{The gender gap in the credit evaluation TPR (female's TPR minus male's TPR). New users only.}
\label{tab:xgb-gap-new}
\end{table}

\begin{table}[H]
\centering
\begin{tabular}{lll}
\hline
           & $\beta_M=-0.1042$                      & $\beta_M=0$ \\ \hline
$c_{iM}=0.2390$  & {2.05\%} & 2.31\% \\ \hline
$c_{iM}=0$ & {2.30\%} & 1.94\% \\ \hline
\end{tabular}
\vskip 3mm
\caption{The gender gap in the credit evaluation TPR (female's TPR minus male's TPR). Repeated users only.}
\label{tab:xgb-gap-repeated}
\end{table}

Similarly, we separate the new users and the repeated users to check the lower level effects (Table~\ref{tab:xgb-gap-new} and Table~\ref{tab:xgb-gap-repeated}). Consistent with our observation in Section~\ref{sec:policy}, repeated users have smaller bias than new users. This indicates that signals can help mitigate bias in machine learning models as well. Although for repeated users, there is a slight increase when removing any one of the two bias (2.05\% $\rightarrow$ 2.30\%, 2.05\% $\rightarrow$ 2.31\%), the increase is not significant (p-value = 0.5312; p-value = 0.3047). This support that for machine learning models, removing bias can help mitigate the bias for both new users and repeated users. When we compare the de-biased outcome of human decision making and algorithm decision making, we find that human can achieve smaller TPR gap (2.04\%) than the algorithm (3.98\%) for new applicants, but lager TPR gap (2.70\%) than the algorithm (1.94\%) for repeated applicants. This finding suggests that the optimal strategy is to provide enough training to human evaluators to eliminate their decision bias, and then use human evaluators on new users while use machine learning algorithm to evaluate repeated users.



\section{Discussion and Conclusions}

We have proposed a framework to use structural modeling to distinguish and estimate two types of human bias, i.e., belief-based bias and preference-based bias, based on observational data. In our micro-lending context, the evaluators hold a persistent preference-based bias but learn from three distinct signals (the final overdue days $D_{it}$, the proportion of installments with positive attitude from the borrower $A_{it}$, the financial help from family and friends $H_{it}$), which updates the evaluators’ belief-based bias. The model was estimated on real-world data, and our model explains the data well.

The estimation results by the structural model imply that the evaluators possess a preference-based bias in favor of female applicants and against male ones; they also hold a belief-based bias with a higher prior belief of females’ credit qualities. By observing the repayment behaviors, the evaluators can quickly update their belief of the borrowers’ credit qualities. And all the three signals play significant roles in the evaluators’ learning. 

The results from our policy simulations suggest that both the eliminations of the preference-based bias and the belief-based bias can increase the platform’s profits. The underlying mechanisms of the two counterfactual settings are the same. Because the loss from lowering the approval probability for nondefault users is smaller than the gain from lowering the approval probability for default users, the platform achieves higher profits. One the borrower side, the eliminations of both types of bias can reduce the gender gap in the credit evaluation true positive rate.

We also feed the real-word dataset and the counterfactual datasets into XGBoost model, to examine how ML algorithms inherit human bias. We find that machine learning algorithm can mitigate both the preference-based bias and the belief-based bias.

Our paper also has certain limitations that can be addressed in future work. First, the microloan users are generally not stable in their financial condition, which may be one plausible reason why the evaluators heavily rely on the latest repayment behaviors to form a belief of borrowers’ credit qualities. In a more stable setting like credit card or mortgage, evaluators may gradually update their beliefs of borrowers’ credit qualities. Second, in our policy simulations, we only consider the changes on the evaluator side. In reality, the changes in previous evaluator approval behaviors can also lead to changes in subsequent application behaviors of borrowers. Future work may take both sides into consideration. Third, as a pioneer work on quantifying different types of bias, we do not consider the interaction of gender and other attributes due to model complexity and identification issues. Future work may explore those interaction effects. Despite these limitations, to our best knowledge, this paper is the first to use structural modeling to uncover and distinguish the different types of bias in decision-making processes based on observational data. As machine learning and AI models are increasingly deployed across many decision-making scenarios, it is more and more important to understand the source of biases and propose well-targeted solutions.

\bibliographystyle{ACM-Reference-Format}
\bibliography{refs}

\end{document}